\documentclass[final]{cvpr}

\usepackage{times}
\usepackage{epsfig}
\usepackage{graphicx}
\usepackage{amsmath}
\usepackage{amssymb}
\usepackage{bm}
\usepackage{wrapfig}
\usepackage{multirow}
\usepackage{placeins}
\usepackage{multirow}
\usepackage{booktabs}

\usepackage[pagebackref=true,breaklinks=true,colorlinks,bookmarks=false]{hyperref}

\def\cvprPaperID{6641} 
\def\confYear{CVPR 2021}

\begin{document}

\setlength{\abovedisplayskip}{2pt}
\setlength{\belowdisplayskip}{2pt}

\title{Causal Attention for Vision-Language Tasks}

\author{Xu Yang$^1$, Hanwang Zhang$^1$, Guojun Qi$^{2}$, Jianfei Cai$^{3}$ \\
$^1$School of Computer Science and Engineering, Nanyang Technological University, Singapore,\\
$^{2}$Futurewei Technologies \\
$^3$Faculty of Information Technology, Monash University, Australia,\\
\tt\small{ s170018@e.ntu.edu.sg, hanwangzhang@ntu.edu.sg, guojunq@gmail.com, Jianfei.Cai@monash.edu}
}

\maketitle

\begin{abstract}
We present a novel attention mechanism: Causal Attention (CATT), to remove the ever-elusive confounding effect in existing attention-based vision-language models. This effect causes harmful bias that misleads the attention module to focus on the spurious correlations in training data, damaging the model generalization. As the confounder is unobserved in general, we use the front-door adjustment to realize the causal intervention, which does not require any knowledge on the confounder. Specifically, CATT is implemented as a combination of 1) In-Sample Attention (IS-ATT) and 2) Cross-Sample Attention (CS-ATT), where the latter forcibly brings other samples into every IS-ATT, mimicking the causal intervention. CATT abides by the Q-K-V convention and hence can replace any attention module such as top-down attention and self-attention in Transformers. CATT improves various popular attention-based vision-language models by considerable margins. In particular, we show that CATT has great potential in large-scale pre-training, e.g., it can promote the lighter LXMERT~\cite{tan2019lxmert}, which uses fewer data and less computational power, comparable to the heavier UNITER~\cite{chen2020uniter}. Code is published in \url{https://github.com/yangxuntu/catt}.
\end{abstract}

\section{Introduction}
Stemming from the strong cognitive evidences in selective signal processing~\cite{tootell1998retinotopy, rensink2000dynamic}, the attention mechanism has arguably become the most indispensable module in vision and language models~\cite{xu2015show, bahdanau2014neural,anderson2018bottom,devlin2018bert,carion2020end,luo2018discriminability}. Although its idiosyncratic formulation varies from task to task, its nature can be summarized as the following common Q-K-V notation: given a \emph{query} $\mathbf{q}$, the attention mechanism associates $\mathbf{q}$ to each feature \emph{value} $\mathbf{v}_i$ by using the normalized attentive weight $\alpha_i\propto \mathbf{q}^T\mathbf{k}_i$, where $\mathbf{k}_i$ is the \emph{key} function of the value; thus, the resultant selective feature value --- attention --- is $\sum_i\alpha_i\mathbf{v}_i$. In a modern view, the attention can be understood as a feature \emph{transformer} that encodes input query $\mathbf{q}$ by using the given values $\bm{V} = \{\mathbf{v}_i\}$~\cite{vaswani2017attention}. 
\begin{figure}[t]
\centering
\includegraphics[width=1\linewidth,trim = 5mm 5mm 5mm 5mm,clip]{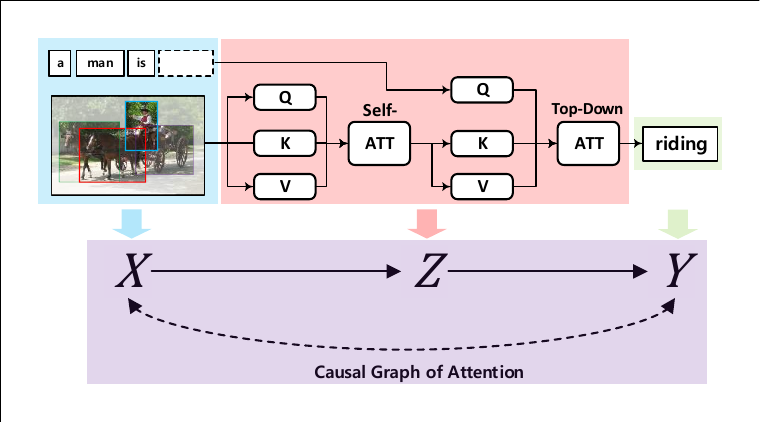}
   \caption{Top: an example of image captioner with a self-attention and a top-down attention modules. Bottom: the corresponding causal graph. The reason why the prediction is ``riding'' but not ``driving'' is explained in Figure~\ref{fig:causal_graph}.
   }
\label{fig:1}
\vspace{-0.3in}
\end{figure}

Taking image captioning as an example in Figure~\ref{fig:1}, if $\mathbf{q}$ and $\bm{V}$ are both encoded from the input $X$, \eg, the RoI features of an image, we call it \emph{self-attention}; if $\mathbf{q}$ is changed to the sentence context, we call it \emph{top-down attention}. Intuitively, self-attention is usually viewed as a non-local~\cite{wang2018non} (or graph~\cite{battaglia2018relational}) convolution network that enriches each local value with global relationship features; top down-attention is used to enrich the context with the cross-domain relationship features~\cite{anderson2018bottom}. Both of them can be combined and stacked into deep networks, serving as powerful multi-modal encoder-decoder transformer networks~\cite{lu2019vilbert,chen2020generative}.

\begin{figure}[t]
\centering
\includegraphics[width=1\linewidth,trim = 5mm 5mm 5mm 5mm,clip]{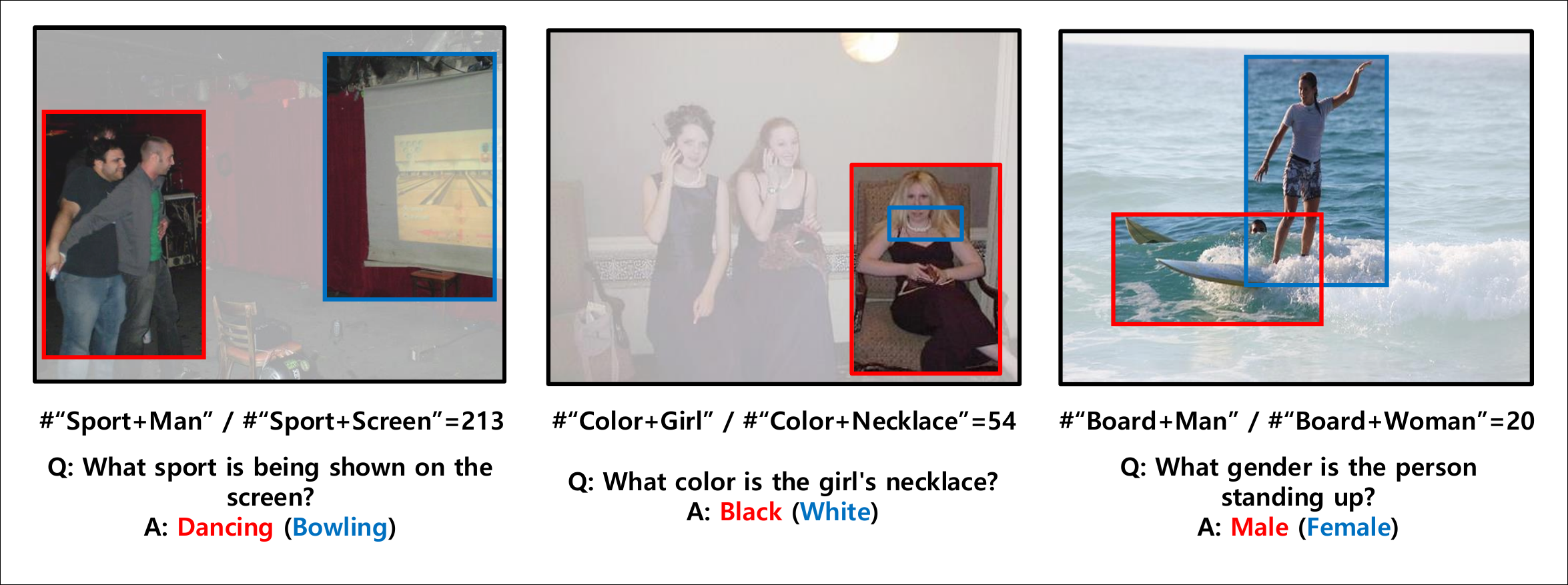}
   \caption{Before pre-training (\eg, LXMERT~\cite{tan2019lxmert}), attentions are correct (blue). After pre-training, attentions are wrong (red). This is because the co-occurrences of some concepts appear much more often than others, \eg, ``Sport+Man'' appears 213 times more than ``Sport+Screen'' in the pre-training data.
   }
\label{fig:bias}
\vspace{-0.3in}
\end{figure}

As a bridge connecting the input feature $X$ and the output label $Y$, the quality of attention --- how reasonable the attentive weight $\alpha$ is --- plays a crucial role for the overall performance. However, due to the fact that the attention weights are unsupervised, \eg, there is no word-region grounding for the top-down attention or relationship dependency annotation for the self-attention, the weights will be inevitably misled by the dataset bias. For example, as shown in Figure~\ref{fig:1}, since there are many images captioned with ``person riding horse'' in the training data, self-attention learns to infer ``riding'' by building the dependency between ``person'' and ``horse''. Then, given a test image with ``person driving carriage'', this self-attention still tends to relate ``person'' with ``horse'' to infer ``riding'', but ignoring the ``carriage''. Unfortunately, such bias cannot be mitigated by simply enlarging the dataset scale, as most of the bias abides by the data nature --- Zipf's law~\cite{reed2001pareto} and social conventions~\cite{hendricks2018women} --- there are indeed more ``red apple'' than ``green apple'' or ``person standing'' than ``person dancing''. Therefore, as shown in Figure~\ref{fig:bias}, large-scale pre-training may lead to even worse attentions.

\begin{wrapfigure}{r}{0.18\textwidth}
    \centering
    \includegraphics[width=1\linewidth,trim = 5mm 5mm 5mm 5mm,clip]{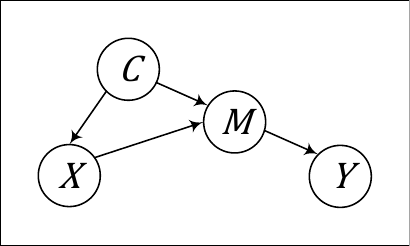}
    \caption{This expands the causal links of the confounding path $X\dashleftarrow\dashrightarrow Y$ in Figure~\ref{fig:1}}.
    \label{fig:causal_graph}
\end{wrapfigure}
The dataset bias is essentially caused by the confounder, a common cause that makes $X$ and $Y$ correlated even if $X$ and $Y$ have no direct causation. We illustrate this crucial idea in Figure~\ref{fig:causal_graph}. Suppose that the confounder $C$ is the common sense\footnote[1]{It is also well-known as the disentangled causal mechanism~\cite{suter2019robustly}.} ``person can ride horse'', $C\rightarrow X$ denotes that a visual scene is generated by such knowledge, \eg, the dataset curator observes and captures the common sense; $X\rightarrow M$ denotes the fact that the objects $M = \{\text{person}, \text{horse}\}$ can be detected (\eg, Faster R-CNN~\cite{ren2015faster}), whose object inventory is determined by $C\rightarrow M$; $M\rightarrow Y$ denotes the language generation for ``person riding horse''. Note that besides the legitimate causal path from image $X$ via object $M$ to $Y$, the ``backdoor'' path $X\leftarrow C\rightarrow M\rightarrow Y$ also contributes an effect to $Y$. Therefore, if we only train the model based on the correlation $P(Y|X)$ without knowing the confounding effect, no matter how large the amount of training data is, the model can never identify the true causal effect from $X$ to $Y$~\cite{pearl2000causality,rubin2005causal}. For example, if the confounder distribution varies from training to testing, \eg, the common sense ``person can ride horse'' is dominantly more often than the common sense ``person can drive carriage'' in training, but the latter is more often than the former in testing, then $P(Y|X)$ based on ``person can ride horse'' in training will be no longer applicable in testing~\cite{pearl2014external}.

In this paper, we propose a novel attention mechanism called: \emph{Causal Attention} (CATT), which can help the models identify the causal effect between $X$ and $Y$, and thus mitigates the bias caused by confounders. It is based on the \emph{front-door adjustment} principle that does not require the assumption of any observed confounder~\cite{pearl1995causal}, and thus CATT can be applied in any domain where the attention resides. In this way, CATT is fundamentally different from existing deconfounding methods based on the backdoor adjustment~\cite{zhang2020causal, wang2020visual}, which has to be domain-specific to comply with the observed-confounder assumption. Specifically, we first show that the conventional attention is indeed an improper approximation of the front-door principle, and then we show what is a proper one, which underpins CATT theoretically  (Section~\ref{subsec:att_vs_frontdoor}). 

We build CATT on the proposed In-Sample attention (IS-ATT) and Cross-Sample attention (CS-ATT), which abides by the Q-K-V operations (Section~\ref{subsec:isatt_and_csatt}). In particular, the parameters of the Q-K-V operations can also be shared between both IS-ATT and CS-ATT to further improve the efficiency in some architectures. We replace the the conventional attention with CATT in various vision-language models to validate its effectiveness, including the classic Bottom-Up Top-Down LSTM~\cite{anderson2018bottom}, Transformer~\cite{vaswani2017attention}, and a large-scale vision-language pre-training (VLP) model LXMERT~\cite{tan2019lxmert}. The experimental results demonstrate that our CATT can achieve consistent improvements for all of them. Significantly, our light LXMERT+CATT outperforms the heavy UNITER~\cite{chen2020uniter} on VQA2.0, \ie, 73.04\% vs. 72.91\% on test-std split, and NLVR2, \ie, 76.0\% vs. 75.80\% on test-P split, while we require much fewer pre-training burdens: 624 vs. 882 V100 GPU hours. Such comparisons show that our CATT has great potential in vision-language pre-training (VLP) tasks.

\section{Related Work}
\noindent\textbf{Attention Mechanism.}
Conventional top-down attentions generally include the classic single-guidance fashion~\cite{bahdanau2014neural,xu2015show,yang2016stacked,you2016image} and the co-guidance fashion~\cite{lu2016hierarchical,yu2019deep}. They can be summarized as the query, key, value (Q-K-V) operation that also generalizes to self-attention~\cite{vaswani2017attention,wang2018non}, which even be applied in pure vision tasks such as visual recognition and generation~\cite{carion2020end, chen2020generative}. As the attention weight is unsupervised, it is easily misled by the confounders hidden in the dataset. We exploit the causal inference to propose a novel CATT module to mitigate the confounding effect~\cite{PearlMackenzie18,pearl2000causality}. As our proposed CATT complies with the Q-K-V convention, it has great potential in any model that uses attention.

\noindent\textbf{Vision-Language Pre-Training.}
Inspired by the success of large-scale pre-training for language modeling~\cite{devlin2018bert,radford2018improving}, researchers have developed some multi-modal Transformer-based Vision-Language Pre-training (VLP) models to learn task-agnostic visiolinguistic representations~\cite{lu2019vilbert,tan2019lxmert,li2019visualbert,chen2020uniter,zhou2019unified,li2020oscar,li2021semvlp}. To discover the visiolinguistic relations across domains, a huge amount of data~\cite{sharma2018conceptual,chen2015microsoft,krishna2017visual} are required for VLP. However, just as the language pre-training models tend to learn or even amplify the dataset bias~\cite{kurita2019measuring,nadeem2020stereoset}, these VLP models may also overplay the spurious correlation. We use the proposed CATT to help VLP models confront the bias.

\noindent\textbf{Causal Inference.}
Causality~\cite{pearl2000causality,rubin2005causal} provides researchers new methodologies to design robust measurements~\cite{suter2019robustly}, discover hidden causal structures~\cite{bengio2019meta}, generate counterfactual samples~\cite{tang2020unbiased,abbasnejad2020counterfactual,kocaoglu2017causalgan,yue2021counterfactual}, and confront various biases~\cite{veitch2020adapting,yue2020interventional,zhang2020causal,niu2020counterfactual,hu2021distilling,qi2020two}. These bias removal methods usually assume that the confounder is observable~\cite{yue2020interventional,zhang2020causal} or domain-specific~\cite{hendricks2018women,bolukbasi2016man}. In general, the confounder is unobservable and elusive. Compared with them, we exploit the front-door adjustment~\cite{pearl1995causal} with no observed-confounder assumption to mitigate the dataset bias. To tackle the sampling challenge in the front-door adjustment, we propose two effective approximations called In-Sample Sampling and Cross-Sample Sampling.

\section{Causal Attention}
\subsection{Attention in the Front-Door Causal Graph}
\label{subsec:att_vs_frontdoor}
We retrospect the attention mechanism in a front-door causal graph~\cite{PearlMackenzie18,pearl2000causality} as shown in the bottom part of Figure~\ref{fig:1}, where the causal effect is passed from the input set $X$ to the target $Y$ through a mediator $Z$. By this graph, we can split the attention mechanism into two parts: a selector which selects suitable knowledge $Z$ from $X$ and a predictor which exploits $Z$ to predict $Y$. Take VQA as the example, $X$ is a multi-modality set containing an image and a question, then the attention system will choose a few regions from the image based on the question to predict the answer. We usually use the observational correlation $P(Y|X)$ as the target to train an attention-based model:
\begin{equation} \label{equ:attention}
\small
   P(Y|X)  =\underbrace{\sum\nolimits_{z}P(Z=z|X)}_{\text{IS-Sampling}}{P(Y|Z=z)},
\end{equation}
where $z$ denotes the selected knowledge and \textbf{IS-Sampling} denotes In-Sample sampling since $z$ comes from the current input sample $X$.

However, as discussed in Introduction, since the selection is an unsupervised process, the predictor may be misled by the dataset bias when training it by Eq.~\eqref{equ:attention}. In causal terms, this means that the predictor may learn the spurious correlation brought by the backdoor path $Z \leftarrow X \leftrightarrow Y$\footnote{For convenience, we simplify the notation of the backdoor path $X \leftarrow C \rightarrow M \rightarrow Y$ shown in Figure~\ref{fig:causal_graph} to $X \leftrightarrow Y$.} instead of the true causal effect $Z \rightarrow Y$, and thus the conventional attention mechanism is not a proper way of calculating the causal effect.

To eliminate the spurious correlation brought by the hidden confounders, we should block the backdoor path between $Z$ and $Y$: $Z \leftarrow X \leftrightarrow Y$. In this way, we can estimate the true causal effect between $Z$ and $Y$, which is denoted as $P(Y|do(Z))$, where $do(\cdot)$ denotes the interventional operation~\cite{pearl2000causality}. We can cut off the link $X \rightarrow Z$ to block this backdoor path by stratifying the input variable $X$ into different cases $\{\bm{x}\}$ and then measuring the average causal effects of $Z$ on $Y$ by the following expectation~\cite{pearl2016causal}:
\begin{equation}\label{equ:front_door3}
\small
    P(Y|do(Z))=\underbrace{\sum\nolimits_{x}P(X=x)}_{\text{CS-Sampling }}P(Y|X=x,Z),
\end{equation}
where $x$ denotes one possible input case. Here we denote it as Cross-Sample Sampling (\textbf{CS-Sampling}) since it comes from the other samples. Intuitively, CS-Sampling approximates the ``physical intervention'' which can break the spurious correlation caused by the hidden confounder. 
For example, the annotation ``man-with-snowboard'' is dominant in captioning dataset~\cite{hendricks2018women} and thus the predictor may learn the spurious correlation between the snowboard region with the word ``man'' without looking at the person region to reason what actually the gender is. CS-Sampling alleviates such spurious correlation by combining the person region with the other objects from other samples, \eg, bike, mirror, or brush, and inputting the combinations to the predictor. Then the predictor will not always see ``man-with-snowboard'' but see ``man'' with the other distinctive objects and thus it will be forced to infer the word ``man'' from the person region. With this deconfounded predictor, the selector will also be forced to select the legitimate evidence even we do not have any region-word supervisions.

By replacing $P(Y|\bm{z})$ in Eq.~\eqref{equ:attention} by $P(Y|do(Z))$ in Eq.~\eqref{equ:front_door3}, we can calculate the true causal effect between $X$ and $Y$:
\begin{equation} \label{equ:front_door}
\small
\begin{aligned}
    &P(Y|do(X))\\
    = &\underbrace{\sum\nolimits_{z}P(Z=z|X)}_{\text{IS-Sampling}}\underbrace{\sum\nolimits_{x}P(X=x)}_{\text{CS-Sampling }}[P(Y|Z=z,X=x)].
\end{aligned}
\end{equation}
This is also called the front-door adjustment, which is a fundamental causal inference technique for deconfounding the unobserved confounder~\cite{pearl1995causal}. Since our novel attention module is designed by using Eq.~\eqref{equ:front_door} as the training target, we name our attention module as \textit{Causal Attention} (CATT). 

\subsection{In-Sample and Cross-Sample Attentions}
\label{subsec:isatt_and_csatt}
To implement our causal attention (Eq.~\eqref{equ:front_door}) in a deep framework, we can parameterize the predictive distribution $P(Y|Z,X)$ as a network $g(\cdot)$ followed by a softmax layer since most vision-language tasks are transformed into classification formulations~\cite{vinyals2015show,antol2015vqa}:
\begin{equation}\label{equ:att_0}
\small
\begin{aligned}
P(Y|Z,X) =\text{Softmax}[g(Z,X)]. \\
\end{aligned}
\end{equation}
As can be seen in Eq.~\eqref{equ:front_door}, we need to sample $X$ and $Z$, and feed them into the network to complete $P(Y|do(X))$. However, the cost of the network forward pass for all of these samples is prohibitively expensive. To address this challenge, we apply Normalized Weighted Geometric Mean (NWGM) approximation~\cite{xu2015show,srivastava2014dropout} to absorb the outer sampling into the feature level and thus only need to forward the ``absorbed input'' in the network for once. Specifically, by NWGM approximation, IS-Sampling and CS-Sampling in Eq.~\eqref{equ:front_door} can be absorbed into the network as:
\begin{equation}\label{equ:att_1}
\small
\begin{aligned}
P(Y|do(X)) \approx & \text{Softmax}[g(\hat{\bm{Z}},\hat{\bm{X}})], \\
\textbf{IS-Sampling:} \quad  \hat{\bm{Z}} =& \sum\nolimits_{z} P(Z=z|h(X)) \bm{z}, \\
\textbf{CS-Sampling:} \quad \hat{\bm{X}} = &\sum\nolimits_{x} P(X=x|f(X)) \bm{x}. 
\end{aligned}
\end{equation}
where $h(\cdot)$ and $f(\cdot)$ denote query embedding functions which can transform the input $X$ into two query sets. Both of them can be parameterized as networks. Note that in a network, the variable $X$ and $Z$ are represented by embedding vectors, \eg, an image region becomes an RoI representation, so we use bold symbols to signify these embedding vectors, \eg, $\bm{z}$, $\bm{x}$ denote the embedding vectors of the variable $z$, $x$. $\hat{\bm{X}}$, $\hat{\bm{Z}}$ denote the estimations of the IS-Sampling and CS-Sampling, which can be packed into the matrix form~\cite{vaswani2017attention}. The derivation details of Eq.~\eqref{equ:att_1} are given in the supplementary material.
 
\begin{figure}[t]
\centering
\includegraphics[width=1\linewidth,trim = 5mm 5mm 5mm 5mm,clip]{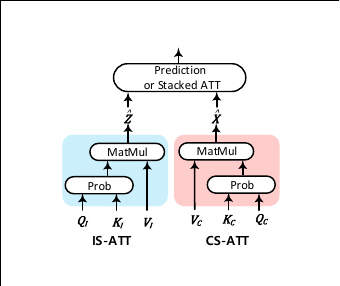}
   \caption{ The sketch of a single causal attention module, which includes an IS-ATT (Eq.~\eqref{equ:isatt}) and a CS-ATT (Eq.~\eqref{equ:csatt}). After calculating $\hat{\bm{Z}}$ and $\hat{\bm{X}}$, we can input them into the predictor for making decisions or more stacked attention layers for further embedding.
   }
\label{fig:catt}
\vspace{-0.2in}
\end{figure}

Actually, the IS-Sampling estimation $\hat{\bm{Z}}$ is what a classic attention network calculates, which can be briefly expressed by the Q-K-V operation as the blue block in Figure~\ref{fig:catt}:
\begin{equation} \label{equ:isatt}
\small
\begin{aligned}
 \textbf{Input:} \quad & \bm{Q}_I, \bm{K}_I, \bm{V}_I,  \\
 \textbf{Prob:} \quad  & \bm{A}_I = \text{Softmax}({\bm{Q}_I}^T {\bm{K}_I}) \\
 \textbf{Ouput:} \quad & \hat{\bm{Z}} = \bm{V}_I\bm{A}_I
\end{aligned}
\end{equation}
We denote Eq.~\eqref{equ:isatt} as the In-Sample attention (\textbf{IS-ATT}) and the subscript ``$I$'' emphasizes that it is estimating IS-Sampling. In this case, all the $\bm{K}_I$ and $\bm{V}_I$ come from the current input sample, \eg, the RoI feature set. $\bm{Q}_I$ comes from $h(X)$, \eg, in top-down attention, the query vector $\bm{q}_I$ is the embedding of the sentence context and in self-attention, the query set $\bm{Q}_I$ is also the RoI feature set. For $\bm{A}_I$, each attention vector $\bm{a}_I$ is the network estimation of the IS-Sampling probability $P(Z=z|h(X))$ and the output $\hat{\bm{Z}}$ is the estimated vector set of IS-Sampling in Eq.~\eqref{equ:att_1}. 

Inspired by Eq~\eqref{equ:isatt}, we can also deploy a Q-K-V operation to estimate $\hat{\bm{X}}$ and name it as Cross-Sample attention (\textbf{CS-ATT}), which is the red block in Figure~\ref{fig:catt}:
\begin{equation} \label{equ:csatt}
\small
\begin{aligned}
 \textbf{Input:} \quad & \bm{Q}_C, \bm{K}_C, \bm{V}_C,  \\
 \textbf{Prob:} \quad  & \bm{A}_C = \text{Softmax}({\bm{Q}_C}^T {\bm{K}_C}), \\
 \textbf{Ouput:} \quad & \hat{\bm{X}} = \bm{V}_C\bm{A}_C
\end{aligned}
\end{equation}
where $\bm{K}_C$, $\bm{V}_C$ come from the other samples in the training set, and $\bm{Q}_C$ comes from $f(X)$. In this case, $\bm{a}_C$ approximates $P(X=x|f(X))$ and $\hat{\bm{X}}$ is the CS-Sampling estimation in Eq.~\eqref{equ:att_1}. In the implementations, we set $\bm{K}_C$ and $\bm{V}_C$ as the global dictionaries compressed from the whole training dataset since it is impossible to attend to all the samples in the training set. Specifically, we initialize this dictionary by using K-means over all the samples' embeddings in training set, \eg, all the images' RoI features. In this way, $\bm{V}_C$ and $\bm{V}_I$ stay in the same representation space, which guarantees that the estimations of IS-Sampling and CS-Sampling: $\hat{\bm{Z}}$ and $\hat{\bm{X}}$ in Eq.~\eqref{equ:att_1} also have the same distribution. 

To sum up, as shown in Figure~\ref{fig:catt}, our single causal attention module estimates $\hat{\bm{Z}}$ and $\hat{\bm{X}}$ respectively by IS-ATT in Eq.~\eqref{equ:isatt} and CS-ATT in Eq.~\eqref{equ:csatt}. After that, we can concatenate the outputs for estimating $P(Y|do(X))$ as in Eq.~\eqref{equ:att_1}. 

\subsection{CATT in Stacked Attention Networks}
In practice, attention modules can be stacked as deep networks, \eg, the classic Transformer~\cite{vaswani2017attention} or BERT architectures~\cite{devlin2018bert}. Our CATT can also be incorporated into these stacked attention networks and we experiment with Transformer~\cite{vaswani2017attention} and LXMERT~\cite{tan2019lxmert} in this paper. We briefly introduce their architectures here and discuss the implementation details in Section~\ref{sec:implementation}. Generally, our CATT replaces the first attention layer of these architectures to get the estimations of IS-Sampling $\hat{\bm{Z}}$ and CS-Sampling $\hat{\bm{X}}$, and then we input them into more attention layers for further embedding, as shown in Figure~\ref{fig:catt}.
For convenience, in these stacked attention networks, we still use IS-ATT and CS-ATT as the names of the attention modules to signify that this attention layer is dealing with the representations of the IS-Sampling or CS-Sampling.
\begin{figure}[t]
\centering
\includegraphics[width=1\linewidth,trim = 5mm 5mm 5mm 5mm,clip]{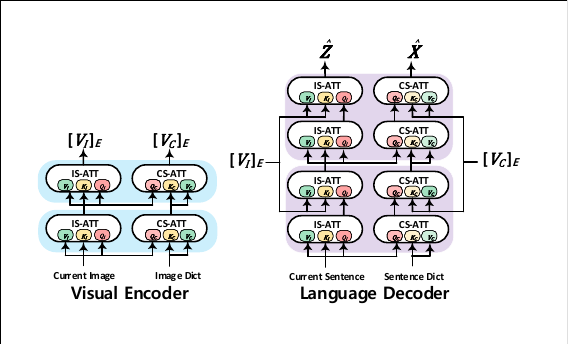}
   \caption{The Transformer+CATT architecture, which contains a visual encoder and a language decoder. We only show two layers in both parts for demonstrating how they are connected. In the implementations, both the encoder and decoder contain 6 layers. $[\bm{V}_I]_E$ and $[\bm{V}_C]_E$ denote the IS-ATT and CS-ATT outputs of the encoder, which are used as the inputs to the decoder. $\hat{\bm{Z}}$ and $\hat{\bm{X}}$ are the IS-ATT and CS-ATT outputs of the decoder, which are the estimations of IS-Sampling and CS-Sampling, respectively.
   }
\label{fig:transformer}
\vspace{-0.2in}
\end{figure}

\noindent\textbf{Transformer+CATT.}
Figure~\ref{fig:transformer} shows the architecture of our vision-language Transformer+CATT. This architecture contains a vision encoder and a language decoder. In implementations, both the encoder and decoder contain 6 blue and purple blocks. The inputs of the encoder include the embedding set of the current image and a global image embedding dictionary. The IS-ATT and CS-ATT outputs of the encoder are input into the decoder for learning visiolinguistic representations. For the decoder, the inputs of the first IS-ATT and CS-ATT are respectively the current sentence embedding set and a global sentence embedding dictionary. The outputs of the decoder include two parts which respectively correspond to IS-Sampling $\hat{\bm{Z}}$ and CS-Sampling $\hat{\bm{X}}$, which will be concatenated and input into the final predictor. Importantly, by stacking many CATT layers, the estimated $\hat{\bm{Z}}$ and $\hat{\bm{X}}$ may not stay in the same representation space due to the non-convex operations in each attention module, \eg, the position-wise feed-forward Networks~\cite{vaswani2017attention}. To avoid this, we share the parameters of IS-ATT and CS-ATT in each CATT and then the outputs of them will always stay in the same representation space, where the detail formations are given in Eq.~\eqref{equ:multi-head}. As a result, the additional attention computation of CATT in LXMERT is $O(K*n)$/$O(n*n)$ at the first/other layer, where $K$ is the size of the global dictionary and $n$ is the number of word/image sequence.

\begin{figure}[t]
\centering
\includegraphics[width=1\linewidth,trim = 5mm 5mm 5mm 5mm,clip]{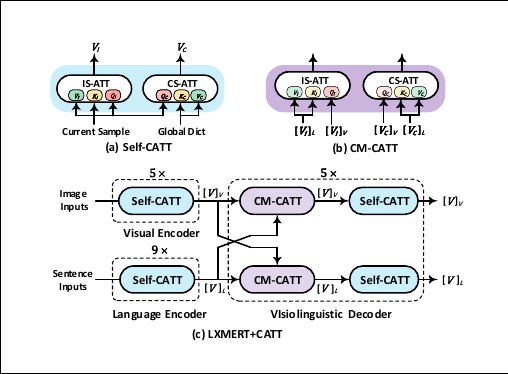}
   \caption{The architecture of LXMERT+CATT, which contains one visual encoder, one language encoder, and one visiolinguistic decoder. Note that each line in (c) contains two parts which respectively correspond to IS-ATT and CS-ATT. $[\bm{V}]_V$ and $[\bm{V}]_L$ denote the visual and language signals, respectively.
   }
\label{fig:lxmert}
\vspace{-0.2in}
\end{figure}

\noindent\textbf{LXMERT+CATT.} Figure~\ref{fig:lxmert} demonstrates the architecture of our LXMERT+CATT, which contains three parts, a vision encoder with 5 self-CATT modules, a language encoder with 9 self-CATT modules, and a visiolinguistic decoder with 5 blocks where each one contains two cross-modality CATT (CM-CATT) and two self-CATT modules. For convenience, we merge the inputs (outputs) of IS-ATT and CS-ATT into one single line in (c). For example, the image inputs contain two parts which are the current image and a global image embedding dictionary. $[\bm{V}]_V$ ($[\bm{V}]_L$) denotes the visual (language) signal which also includes two parts $[\bm{V}_I]_V$ and $[\bm{V}_C]_V$ ($[\bm{V}_I]_L$ and $[\bm{V}_C]_L$) corresponding to IS-ATT and CS-ATT, respectively. Figure~\ref{fig:lxmert}(b) sketches one cross-modality module used in the top part of the decoder in (c), where the visual signals are used as the queries in both IS-ATT and CS-ATT. Similar as the original LXMERT~\cite{tan2019lxmert}, we concatenate the outputs of both vision and language streams and input them into various predictors for solving different vision-language tasks. In implementations, we share the parameters of IS-ATT and CS-ATT in each causal attention module to force their outputs to have the same distributions.

\section{Experiments}
\label{sec:experiments}
We validated our Causal Attention (CATT) in three architectures for various vision-language tasks: Bottom-Up Top-Down (BUTD) LSTM~\cite{anderson2018bottom} for Image Captioning (IC)~\cite{chen2015microsoft,luo2017image} and Visual Question Answering (VQA)~\cite{antol2015vqa}, Transformer~\cite{vaswani2017attention} for IC and VQA, and a large scale vision-language pre-training framework LXMERT~\cite{tan2019lxmert} for VQA, Graph Question Answering (GQA)~\cite{hudson2019gqa}, and Natural Language for Visual Reasoning (NLVR)~\cite{suhr2018corpus}.

\subsection{Datasets}
\noindent\textbf{MS COCO~\cite{chen2015microsoft}} has 123,287 images and each image is assigned with 5 captions. This dataset has two popular splits: the Karpathy split~\cite{karpathy2015deep} and the official test split, which divide the whole dataset into $113,287/5,000/5,000$ and $82,783/40,504/40,775$ for training/validation/test, respectively. We used the Karpathy split to train the BUTD and Transformer based captioners and evaluate.

\noindent\textbf{VQA2.0~\cite{goyal2017making}} collects the images from MS COCO and assigns 3 questions for each image and 10 answers for each question. There are 80k/40k training/validation images available offline. We exploited the training set to train our BUTD and Transformer based VQA systems, and then evaluated the performances on three different splits: offline validation, online test-development, and online test-standard.

\noindent\textbf{Pre-training and Fine-tuning Datasets for VLP.} We followed LXMERT~\cite{tan2019lxmert} to collect a large-scale vision-language pre-training dataset from the training and development sets of MS COCO, VQA2.0, GQA~\cite{hudson2019gqa}, and Visual Genome~\cite{krishna2017visual}. After collecting, this dataset contained 180K distinct images and 9.18M image-sentence pairs. We fine-tuned our VLP model on three tasks, which were VQA, GQA, and NLVR2~\cite{suhr2018corpus} and evaluated the performances on various test splits of them.

\subsection{Implementation Details}
\label{sec:implementation}
The implementation details of BUTD+CATT and Transformer+CATT are given in C.4 of the supplementary material. Here we provide the details of  LXMERT+CATT, which is the most significant experiments in this paper.

\noindent\textbf{LXMERT + CATT.}
We used the architecture in Figure.~\ref{fig:lxmert} for large-scale vision-language pre-training. In this architecture, all IS-ATT and CS-ATT were deployed by 12-head scaled dot-product~\cite{vaswani2017attention}:
\begin{equation} \label{equ:multi-head}
\small
\begin{aligned}
 \textbf{Input:} \quad  &\bm{Q},\bm{K},\bm{V} \\
 \textbf{Prob:} \quad  &\bm{A}_i=\text{Softmax}( \frac{\bm{Q}\bm{W}_i^Q(\bm{K}\bm{W}_i^K)^T}{\sqrt{d}} ) \\
 \textbf{Single-Head}:  \quad  &\bm{H}_i=\bm{A}_i\bm{V}\bm{W}_i^V,\\
 \textbf{Ouput:} \quad & \hat{\bm{V}}= \text{Embed}([\bm{H}_1,...,\bm{H}_12]\bm{W}^H),
\end{aligned}
\end{equation}
where $\bm{W}_i^{*}$ and $\bm{W}^{H}$ are all trainable matrices; $\bm{A}_i$ is the soft attention matrix for the $i$-th head;  $[\cdot]$ denotes the concatenation operation, and Embed$(\cdot)$ means the feed-forward network and the residual operation as in~\cite{vaswani2017attention}. The hidden size was set to 768. Importantly, we shared the parameters between IS-ATT and CS-ATT in each CATT to make the outputs stay in the same representation space. In this case, we also applied K-means to get the initializations and set the size of both dictionaries to 500. We extracted 36 RoI object features from each image by a Faster-RCNN~\cite{ren2015faster} pre-trained on VG as in~\cite{anderson2018bottom}. 

We followed the original LXMERT~\cite{tan2019lxmert} to pre-train our LXMERT+CATT architecture by four tasks: masked cross-modality language modeling, masked object prediction, cross-modality image sentence matching, and image question answering. We used Adam optimizer with a linear-decayed learning rate schedule~\cite{devlin2018bert} where the peak learning rate was set to $5e^{-5}$. We pre-trained the model 20 epochs on 4 GTX 1080 Ti with a batch size of 192. The pre-training cost 10 days. To fairly compare the pre-training GPU hours with UNITER~\cite{chen2020uniter}, we also carried an experiment by 4 V100 with batch size as 256 and it cost 6.5 days for pre-training. When fine-tuning the pre-trained model on VQA2.0, GQA, and NLVR2, the batch size was 32, training epochs was 4, and the learning rates were set to $5e^{-5}$, $ 1e^{-6}$, and $5e^{-5}$, respectively. 


\subsection{Results and Analysis.}
\subsubsection{Image Captioning (IC)}
\begin{table}[t]
\begin{center}
\caption{The performances of various captioners on Karpathy split.}
\label{table:tab_ic}
\scalebox{0.9}{
\begin{tabular}{l c c c c c}
		\hline
		   Models & B@4& M & R &  C & S\\ \hline
		   BUTD~\cite{anderson2018bottom} & $37.2$& $27.5$ & $57.3$ & $125.3$ & $21.1$\\
		   LBPF~\cite{qin2019look}  & $38.3$ & $\bm{28.5}$ & $58.4$ & $127.6$ & $22.0$ \\
		   GCN-LSTM~\cite{yao2018exploring}  & $38.2$ & $\bm{28.5}$ & $58.3$ & $127.6$ & $22.0$ \\
           SGAE~\cite{yang2019auto} & $38.4$ & $28.4$ & $\bm{58.6}$ & $127.8$ & $\bm{22.1}$ \\ 
		   BUTD+CATT & $\bm{38.6}$& $\bm{28.5}$ & $\bm{58.6}$ & $\bm{128.3}$ & $21.9$\\          
		   \hline
		   Transformer       & $38.6$ & $28.5$ & $58.4$ & $128.5$ & $22.0$ \\ 
           VLP~\cite{zhou2019unified}  & $39.5$ & $-$ & $-$ & $129.3$ & $\bm{23.2}$ \\ 
           AoANet~\cite{huang2019attention} & $38.9$ & $29.2$ & $58.8$ & $129.8$ & $22.4$\\ 
           $\mathcal{M}^2$Transformer~\cite{cornia2020meshed} & $39.1$ & $29.2$ & $58.6$ & $131.2$ & $22.6$ \\ 
           
           Transformer+CATT  & $39.4$ & $\bm{29.3}$ & $\bm{58.9}$ & $\bm{131.7}$ & $22.8$ \\
           \hline 
\end{tabular}
}
\end{center}
\vspace{-0.2in}
\end{table}
\noindent\textbf{Similarity Measurements.}
The results are reported in Table~\ref{table:tab_ic}, where the top and bottom parts list various models which respectively deploy LSTM and Transformer as the backbones. In this table, B, M, R, C, and S denote BLEU~\cite{papineni2002bleu}, METEOR\cite{banerjee2005meteor}, ROUGE~\cite{lin2004rouge}, CIDEr-D~\cite{vedantam2015cider}, and SPICE~\cite{anderson2016spice}, respectively, which evaluate the similarities between the generated and the ground-truth captions. 

Compared with two baselines BUTD and Transformer, we can find that BUTD+CATT and Transformer+CATT respectively achieve 3.0-point and 3.2-point improvements on CIDEr-D. More importantly, after incorporating our CATT modules into BUTD and Transformer, they have higher CIDEr-D scores than certain state-of-the-art captioners which deploy more complex techniques. For example, SGAE exploits scene graphs to transfer language inductive bias or $\mathcal{M}^2$Transformer learns multi-level visual relations though additional meshed memory networks. These comparisons suggest that our CATT module is a more powerful technique compared with the techniques used in these state-of-the-art captioners.

\begin{table}[t]
\begin{center}
\caption{The bias degree of different models: ``$\uparrow$'' and ``$\downarrow$'' mean the higher the better and the lower the better, respectively. Red numbers denote the improvements after using our CATT modules.}
\label{table:tab_ic_bias}
\scalebox{0.7}{
\begin{tabular}{l c c c c c}
		\hline
		   Models   & CHs$\downarrow$ & CHi$\downarrow$ & A@Gen$\uparrow$ & A@Attr$\uparrow$ & A@Act$\uparrow$\\ \hline
           BUTD             & 13.5 & 8.9 & 77\% & 41\% & 52\% \\ 
           BUTD+CATT        & 10.7$_{\color{red}{-2.8}}$ & 7.2$_{\color{red}{-1.7}}$ & 85\%$_{\color{red}{+8\%}}$ & 51\%$_{\color{red}{+10\%}}$ & 60\%$_{\color{red}{+8\%}}$ \\ 
           \hline
           Transformer      & 12.1 & 8.1 & 82\% & 47\% & 55\% \\ 
           Transformer+CATT & $9.7_{\color{red}{-2.4}}$ & $6.5_{\color{red}{-1.6}}$ & $92\%_{\color{red}{+10\%}}$ & $56\%_{\color{red}{+9\%}}$ & $64\%_{\color{red}{+9\%}}$ \\ 
           \hline 
\end{tabular}
}
\end{center}
\vspace{-0.3in}
\end{table}
\begin{figure}[t]
\centering
\includegraphics[width=1\linewidth,trim = 5mm 5mm 5mm 5mm,clip]{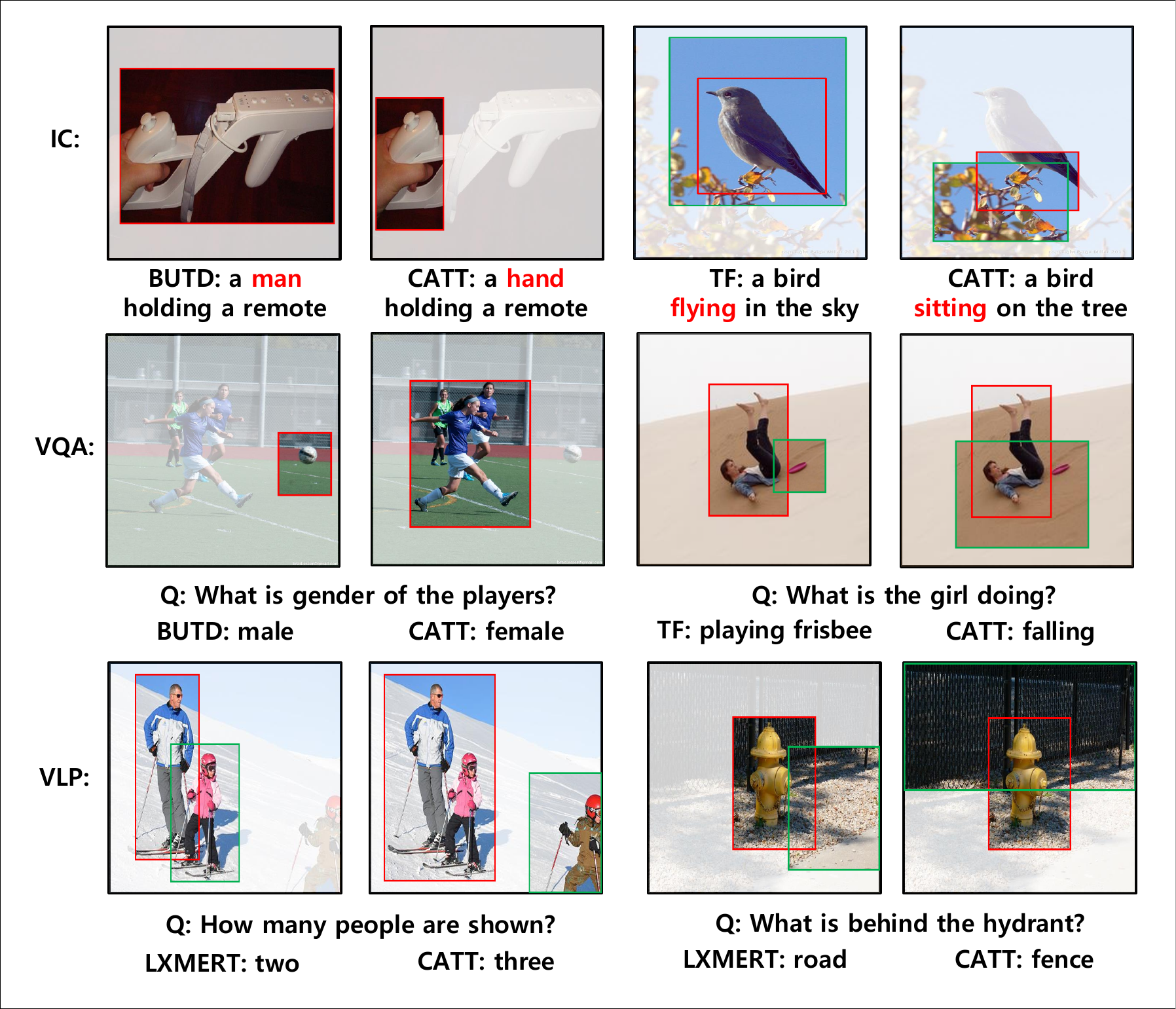}
  \caption{Six examples show that our CATT can correct different models to focus on the suitable regions. TF denotes Transformer. For BUTD, we show the region with the highest attention weight. For Transformer and VLP, the red region has the highest attention weight in top-down attention and the green region is the one most related to the red region in self-attention.
  }
\label{fig:exp}
\end{figure}
\begin{table}[t]
\begin{center}
\caption{Accuracy of various VQA models on different splits.}
\label{table:tab_vqa}
\scalebox{1}{
\begin{tabular}{l c c c }
		\hline  
		   Models& loc-val & test-dev & test-std \\ \hline
		  BUTD~\cite{anderson2018bottom} & 63.20 & 65.32  & 65.67   \\ 
          MUTAN~\cite{ben2017mutan}      & - & 66.01  & 66.38   \\ 
          MLB~\cite{kim2016hadamard}     & 65.07 & 66.27  & 66.62   \\ 
          BUTD+CATT    & $\bm{65.12}$ & $\bm{67.13}$& $\bm{67.26}$  \\
          \hline 
          Transformer & 66.29 & 69.53 & 69.82 \\
          DFAF~\cite{gao2019dynamic}  & 66.21 &  70.22  & 70.34  \\
          MCAN~\cite{yu2019deep} & 67.20 &  70.63  & 70.90  \\
          TRRNet~\cite{yang2020trrnet} & -& 70.80 & 71.20  \\
          Transformer+CATT & $\bm{67.33}$ & $\bm{70.95}$  & $\bm{71.27}$  \\
          \hline
\end{tabular}
}
\end{center}
\vspace{-0.3in}
\end{table}
\begin{table}[t]
\begin{center}
\caption{Accuracy of different question types on test-std split. Red numbers denote the improvements after using our CATT modules.}
\label{table:tab_vqa_bias}
\scalebox{0.9}{
\begin{tabular}{l c c c}
		\hline
		   Models   & \textit{Yes/No} & \textit{Number} &  \textit{Other} \\ \hline
          BUTD               &  81.82 & 44.21 & 56.05 \\ 
          BUTD+CATT          &  83.42$_{\color{red}{+1.6}}$ & 48.96$_{\color{red}{+4.75}}$ & 57.3$_{\color{red}{+1.25}}$  \\ 
          \hline
          Transformer        &  86.25 & 50.7 & 59.9\\ 
          Transformer+CATT   &  87.40$_{\color{red}{+1.15}}$ & 53.45$_{\color{red}{+2.75}}$ & 61.3$_{\color{red}{+1.4}}$   \\ 
          \hline
          LXMERT${^\dagger}$         & 88.17 & 52.63 & 62.73 \\
          LXMERT+CATT  & 88.6$_{\color{red}{+0.43}}$ & 55.48$_{\color{red}{+2.85}}$ & 63.39$_{\color{red}{+0.66}}$ \\
          \hline 
\end{tabular}
}
\end{center}
\vspace{-0.4in}
\end{table}
\begin{table*}[t]
\begin{center}
\caption{Training burdens and performances of different large-scale vision-language pre-training models. ``M'' denotes million.}
\label{table:tab_vlp}
\scalebox{0.9}{
\begin{tabular}{l c c c c c c c c}
		\hline
		    \multirow{2}*{Models} & \multicolumn{2}{c}{Training Burdens}  &  \multicolumn{2}{c}{VQA2.0} & \multicolumn{2}{c}{GQA}  & \multicolumn{2}{c}{NLVR2 (Pair)}\\ 
		    \cmidrule(r){2-3} \cmidrule(r){4-5} \cmidrule(r){6-7} \cmidrule(r){8-9} 
		    & GPU Hours &Image / Text& test-dev & test-std & test-dev & test-std & loc-val  & test-P \\
		    \hline
		  LXMERT~\cite{tan2019lxmert}  & 960 (titan xp) &0.18M / 9.18M  & 72.42  &  72.54 & 60.00 & 60.30 &74.9& 74.5 \\ 
		  LXMERT${^\dagger}$~\cite{tan2019lxmert} & 816 (1080Ti) & 0.18M / 9.18M & 71.96  &  72.18 & 59.90 & 59.94 &74.8& 74.4 \\ 
		  ERNIE-VIL~\cite{yu2020ernie}  & - & 4.20M / 9.58M & 72.62 & 72.85& - & - & - & -  \\
		  UNITER~\cite{chen2020uniter} & 882 (V100) & 4.20M / 9.58M& 72.80  &  72.91& - & - &75.85& 75.80  \\
		  12IN1~\cite{lu202012} & 960 (V100) &  5.40M / 7.48M & -  &  72.92  & 60.48 & - & - & -  \\ 
		  LXMERT${^\dagger}$+CATT & 960 (1080Ti), 624 (V100) &0.18M / 9.18M & $72.81$ & $73.04$ & $60.84$ & $61.17$ &$76.40$& $76.00$ \\ 
		  LXMERT${^\dagger}$+CATT$\uparrow$ & 1536 (1080Ti), 1056 (V100) &0.18M / 9.18M & $\bm{73.54}$ & $\bm{73.63}$ & $\bm{61.87}$ & $\bm{62.07}$ &$\bm{77.27}$& $\bm{77.23}$ \\ 
		  \hline
           
\end{tabular}
}
\end{center}
\vspace{-0.4in}
\end{table*}
\noindent\textbf{Bias Measurements.}
We measured the bias degree of the generated captions in Table~\ref{table:tab_ic_bias} to validate that whether our CATT module can alleviate the dataset bias or not. In this table, CHs and CHi denote CHAIR$_s$ and CHAIR$_i$~\cite{rohrbach2018object}, which are designed to measure the object bias. Apart from them, we also analyze three more specific biases: gender bias, action bias, and attribute bias by calculating the accuracy of these words, which are denoted as A@Gen, A@Attr, and A@Act, respectively. From the results, we can see that after incorporating our CATT module, both BUTD and Transformer generate less biased captions, \eg, the accuracies of gender, attribute, and action are respectively improved by 10\%, 9\%, and 9\% when CATT is used in Transformer. The first row of Figure~\ref{fig:exp} shows two examples where BUTD and Transformer respectively attend to unsuitable regions and then generate incorrect captions, \eg, BUTD attend to the remote region and infer the word ``man'' due to the dataset bias, while our CATT corrects this by attending to the hand region to generate the word ``hand''.

\subsubsection{Visual Question Answering (VQA)}
\label{sec:experiment_vqa}
The top and bottom parts of Table~\ref{table:tab_vqa} respectively report the performances of various LSTM and Transformer based VQA models, where loc-val, test-dev, and test-std denote the offline local validation, online test-development, and online test-standard splits. From this table, we can observe that after deploying our CATT module into BUTD and Transformer, the accuracies are consistently improved. More importantly, the deconfounded BUTD and Transformer outperform certain state-of-the-art models which are better than the original BUTD and Transformer. 

Table~\ref{table:tab_vqa_bias} reports the accuracies of different question types on test-std split. It can be found that the accuracy of \textit{number} has the largest improvements after using CATT modules, \ie, 4.75-point and 2.75-point for BUTD and Transformer, respectively. Significantly, Transformer+CATT has a higher \textit{number} accuracy than the large-scale pre-training model LXMERT: 53.45 vs. 52.63. As analyzed in~\cite{zhang2018learning}, the counting ability depends heavily on the quality of the attention mechanism that a VQA model cannot correctly answer \textit{number} questions without attending to all the queried objects. Thus the consistent improvements in \textit{number} support that our CATT modules can largely ameliorate the quality of the conventional attention mechanism. The second row of Figure~\ref{fig:exp} shows that after incorporating CATT, BUTD and Transformer based VQA models can attend to the right regions for answering the questions.

\subsubsection{Vision-Language Pre-training (VLP)}

           
Table~\ref{table:tab_vlp} shows the training burdens and the performances of various large-scale pre-training models on VQA2.0, GQA, and NLVR2. Note that LXMERT${^\dagger}$ and LXMERT respectively denote the results got from the officially released code and from the published paper. For ERNIE-VIL~\cite{yu2020ernie} and UNITER~\cite{chen2020uniter}, they both have a BASE version and a LARGE version where BASE (LARGE) uses 12 (16) heads and 768 (1024) hidden units in multi-head product operations. We report the performances of their BASE versions since our model used 12 heads and 768 hidden units. For NLVR2, we report the performances of UNITER with the same Pair setting as our model.\footnote{The details of NLVR2 setting can be found in Table 5 of UNITER~\cite{chen2020uniter}.} 

From this table, we can see that compared with LXMERT${^\dagger}$, our LXMERT${^\dagger}$+CATT respectively achieves 0.86, 1.23, 1.6-point improvements on the test-std splits of VQA2.0 and GQA and the test-P split of NLVR2. For example, compared with UNITER which uses fp16, our LXMERT${^\dagger}$+CATT uses fewer GPU hours and the pre-training data, while we have higher performances on VQA2.0: 73.04 vs. 72.91, and NLVR2: 76.0 vs. 75.80. Furthermore, inspired by UNITER~\cite{chen2020uniter}, we enhanced our model and named this one as \textbf{LXMERT+CATT}$\bm{\uparrow}$ by using conditional masking and more RoI features. Specifically, we extracted 64 RoI features from each image to guarantee that our model can be trained on 4 1080 Ti GPUs. It can be found that after using two insights from UNITER, our \textbf{LXMERT+CATT}$\bm{\uparrow}$ can achieve higher performances than UNITER, even though we do not extract 100 RoI features for each image as them. These comparisons suggest that our CATT has great potential in large-scale VLP.

Also, as shown in Table~\ref{table:tab_vqa_bias}, after incorporating CATT into LXMERT, we can observe that the accuracy of \textit{Number} is further improved: 55.48 vs. 52.63, which suggests that our CATT improves the quality of the attention modules in VLP models. The third row of Figure~\ref{fig:exp} shows two examples where CATT modules correct LXMERT to focus on the right regions for answering the questions.

\subsection{Ablation Studies}
We carried exhaustive ablation studies to validate three variants of our causal attention module: K-means initialization, dictionary size, parameter sharing. In particular, we deployed these ablation studies in Transformer+CATT and LXMERT+CATT architectures. 

\noindent\textbf{Comparing Methods.}
\noindent\textbf{Base:} We denote the original Transformer and LXMERT architectures as Base.
\noindent\textbf{CATT w/o Init:} We use CATT to denote the architectures which deploy the CATT modules as in Section~3.3. We did not use the K-means algorithm to initialize the global dictionaries but randomly initialized them. We shared the parameters between IS-ATT and CS-ATT in these models.
\noindent\textbf{CATT w/o Share:} We did not share the parameters between IS-ATT and CS-ATT. Here we used the K-means algorithm to initialize the dictionaries.
\noindent\textbf{CATT+D\#K:} We set the size of the global image and word embedding dictionaries to $K$ by the K-means algorithm and shared the parameters between IS-ATT and CS-ATT.
\begin{table}[t]
\begin{center}
\caption{The performances of various CATT ablation studies on the local validation sets. IC denotes Image Captioning and we report the CIDEr-D score of these IC methods. For the other tasks, we report their accuracies. The left and right parts respectively show the ablation studies of the Transformer and LXMERT architectures. }
\label{table:ablation}
\scalebox{0.80}{
\begin{tabular}{l c c c c c}
		\hline
		    \multirow{2}*{Models} & \multicolumn{2}{c}{Transformer}  &  \multicolumn{3}{c}{LXMERT} \\ 
		    \cmidrule(r){2-3} \cmidrule(r){4-6} 
		    & IC & VQA &VQA& GQA & NLVR2 \\
		    \hline
		  Base           & 128.5 & 66.29 & 69.52 & 59.82 & 74.80 \\ 
		  CATT w/o Init  & 129.8 & 66.56 & 69.81 & 60.14 & 75.22 \\  
		  CATT w/o Share & 130.6 & 66.94 & 70.05 & 60.41 & 75.78 \\ 
		  CATT+D\#100    & 131.1 & 67.02 & 70.12 & 60.62 & 75.94 \\  
		  CATT+D\#200    & 131.4 & 67.21 & 70.29 & 60.77 & 76.26 \\  
		  CATT+D\#500    & $\bm{131.7}$ & $\bm{67.33}$ & $\bm{70.40}$ & $\bm{60.90}$ & $\bm{76.40}$\\  
		  
		  \hline
           
\end{tabular}
}
\end{center}
\vspace{-0.4in}
\end{table}

\noindent\textbf{Results and Analysis.}
Table~\ref{table:ablation} reports the performances of the ablation studies on the local validation sets of different tasks. Firstly, we can observe that after using our CATT architecture, even without K-means initialization or parameter sharing, the performances are better than Base models. Also, we can observe that both K-means initialization and parameter sharing are useful for improving the performances. For example, in LXMERT+CATT, after using K-means and sharing the parameters, the performances of VQA are respectively boosted: 70.40 vs. 69.81 and 70.40 vs. 70.05. Such observation suggests that both strategies encourage the estimated IS-Sampling and CS-Sampling to stay in the same representation space, which is indeed beneficial in improving the performances. Also, by comparing the performances with different dictionary sizes, we can find that bigger dictionaries have better performances.



\section{Conclusion}
In this paper, we exploited the causal inference to analyze why the attention mechanism is easily misled by the dataset bias and then attend to unsuitable regions. We discovered that the attention mechanism is an improper approximation of the front-door adjustment and thus fails to capture the true causal effect between the input and target. Then a novel attention mechanism: causal attention (CATT) was proposed based on the front-door adjustment, which can improve the quality of the attention mechanism by alleviating the ever-elusive confounding effect. Specifically, CATT contains In-Sample and Cross-Sample attentions to estimate In-Sample and Cross-Sample samplings in the front-door adjustment and both of two attention networks abide by the Q-K-V operations. We implemented CATT into various popular attention-based vision-language models and the experimental results demonstrate that it can improve these models by considerable margins. In particular, CATT can promote a light VLP model comparable to a heavy one, which demonstrates its great potential in large-scale pre-training.

This supplementary document will further detail the following aspects in the submitted manuscript: A. Causal Preliminaries, B. Formula Derivations, C. More results D. Implementation Details. 

\section{Causal Preliminaries}
\subsection{Structural Causal Model}
\begin{figure}[t]
\centering
\includegraphics[width=1\linewidth,trim = 5mm 5mm 5mm 5mm,clip]{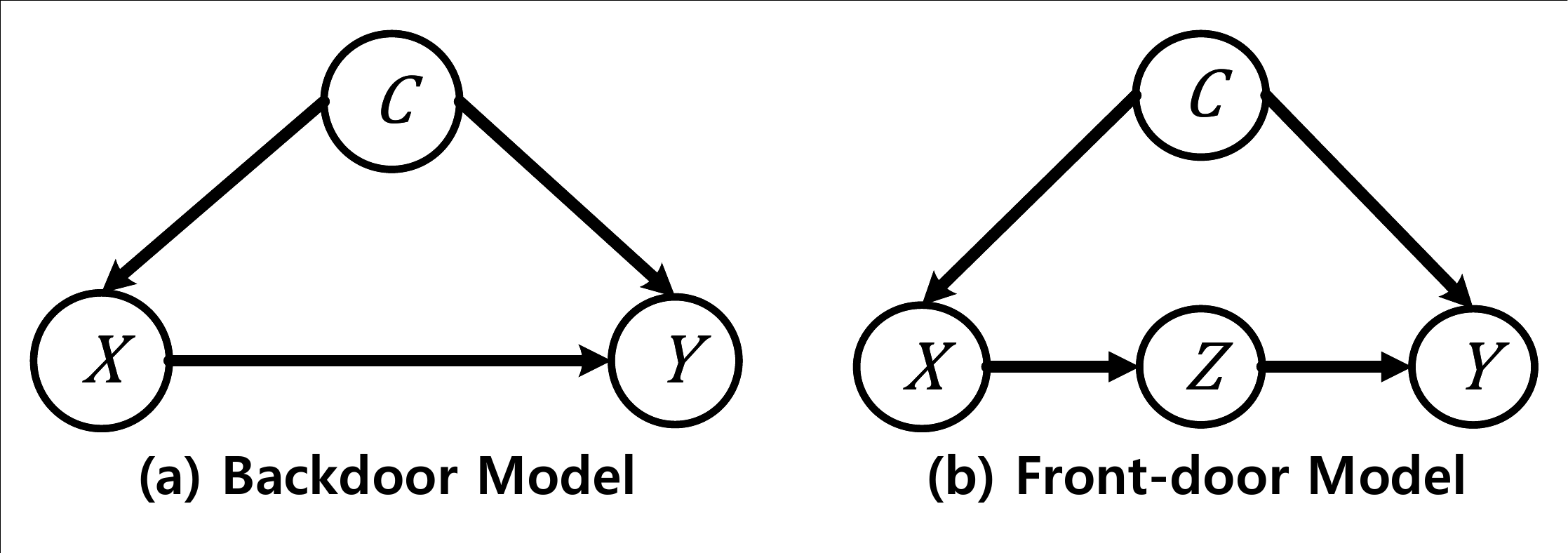}
   \caption{Two Structural Causal Models which are (a) a backdoor model and (b) a front-door model.
   }
\label{fig:fig_supp_cg}
\end{figure}
In Causality~\cite{pearl2016causal,PearlMackenzie18}, a Structural Causal Model (SCM) is used to describe the causal relationships. Such a graph connects different variables by directed edges which denote the causal directions. For example, as shown in Figure~\ref{fig:fig_supp_cg}(a), $X\rightarrow Y$ denotes that $X$ is the cause of $Y$. In an SCM, if a variable is the common cause of two variables, it is called the \textbf{confounder}. For example, $C$ is the cause of both $X$ and $Y$, thus it is a confounder which will induce spurious correlation between $X$ and $Y$ to disturb the recognition of the causal effect between them. In particular, such spurious correlation is brought by the \textbf{backdoor path} created by the confounder. Formally, a backdoor path between $X$ and $Y$ is defined as \textbf{any path from $X$ to $Y$ that starts with an arrow pointing into $X$}. For example, in Figure~\ref{fig:fig_supp_cg}(a), the path $X \leftarrow C \rightarrow Y$ is a backdoor path. Here we use another two examples for helping understand this concept, as in Figure~\ref{fig:fig_supp_cg}(b), $X \leftarrow C \rightarrow Y \leftarrow Z$ and $Z \leftarrow X \leftarrow C \rightarrow Y$ are two backdoor paths between $X$ and $Z$ and $Z$ and $Y$, respectively.

In an SCM, if we want to deconfound two variables $X$ and $Y$ to calculate the true causal effect, we should block every backdoor path between them~\cite{PearlMackenzie18}. For example, in Figure~\ref{fig:fig_supp_cg}(a), we should block $X \leftarrow C \rightarrow Y$ to get the causal effect between $X$ and $Y$.

\subsection{Blocking Three Junctions}
\label{subsec:blocking}
In an SCM, there are three elemental ``junctions'' which construct the whole graph and we have some basic rules to block them. In particular, three junctions are given as follows:

1. $X \rightarrow Z \rightarrow Y$. This is called \textbf{chain junction}, which constructs a front-door path between $X$ and $Y$, as shown in Figure~\ref{fig:fig_supp_cg}(b). In this junction, once we know the value of the mediator $Z$, learning about $X$ will not give us any information to raise or lower our belief about $Y$. Thus, if we know what $Z$ is or directly intervene it as a specific value, we block this chain junction.

2. $X\leftarrow C \rightarrow Y$. This is called \textbf{confounding junction} which induces spurious correlation between $X$ and $Y$, as shown in Figure~\ref{fig:fig_supp_cg}(a). In this junction, once we know what the value of $C$ is or directly intervene it to a specific value, there is no spurious correlation between $X$ and $Y$ and thus we block this junction. 

3. $Z \rightarrow Y \leftarrow C$. This is called \textbf{``collider''} which works in an exactly opposite way from the above chain and confounding junctions. Once we know what the value of $Y$ is, $Z$ and $C$ are correlated. However, if we do not know what $Y$ is or do not intervene it, $Z$ and $C$ are independent and this junction is naturally blocked.

To sum up, if we want to block a path between two variables, we should intervene the middle variables in the chain and confounding junctions and should not intervene in the collider junction. To block a long path, we only need to block a junction of it, \eg, for $X \leftarrow C \rightarrow Y \leftarrow Z$ in Figure~\ref{fig:fig_supp_cg}(b), we can block $X \leftarrow C \rightarrow Y$ by intervening $C$ or block $C \rightarrow Y \leftarrow Z$ by not intervening $Y$.

\subsection{The Backdoor Adjustment}
The backdoor adjustment is the simplest formula to eliminate the spurious correlation by approximating the ``physical intervention''. Formally, it calculates the average causal effect of one variable on another at each stratum of the confounder. For example, in Figure~\ref{fig:fig_supp_cg}(a), we can calculate the causal effect of $X$ on $Y$ as $P(Y|do(X))$:
\begin{equation}
\label{equ:supp_equ_do}
    P(Y|do(X))=\sum\nolimits_c{P(Y|X,C=c)P(C=c)},
\end{equation} 
where $do(\cdot)$ signifies that we are dealing with an active intervention rather than a passive observation. The role of Eq.~\eqref{equ:supp_equ_do} is to guarantee that in each stratum $c$, $X$ is not affected by $C$ and thus the causal effect can be estimated stratum by stratum from the data. 

\subsection{The Front-door Adjustment}
From Eq.~\eqref{equ:supp_equ_do}, we find that to use the backdoor adjustment, we need to know the details of the confounder for splitting it into various strata. However, in our case, we have no idea about what constructs the hidden confounders in the dataset, thus we are unable to deploy the backdoor adjustment. Fortunately, the front-door adjustment~\cite{pearl1995causal} does not require any knowledge on the confounder and can also calculate the causal effect between $X$ and $Y$ in a front-door SCM as in Figure~\ref{fig:fig_supp_cg}(b). 

In Section~3.1 of the submitted manuscript, we have shown the derivation of the front-door adjustment from the attention mechanism perspective. Here we demonstrate a more formally derivation. The front-door adjustment calculates $P(Y|do(X))$ in the front-door $X \rightarrow Z \rightarrow Y$ by chaining together two partially causal effects $P(Z|do(X))$ and $P(Y|do(Z))$: \begin{equation}
\label{equ:supp_fd1}
    P(Y|do(X)) = \sum\nolimits_{z}P(Z=z|do(X))P(Y|do(Z=z)).
\end{equation}

To calculate $P(Z=z|do(X))$, we should block the backdoor path $X \leftarrow C \rightarrow Y \leftarrow Z$ between $X$ and $Z$. As we discussed in Section~\ref{subsec:blocking} that a collider junction is naturally blocked and here $C \rightarrow Y \leftarrow Z$ is a collider, thus this path is already blocked and we have:
\begin{equation}
\label{equ:supp_fd2}
    P(Z=z|do(X)) = P(Z=z|X).
\end{equation}
For $P(Y|do(Z))$, we need to block the backdoor path $Z \leftarrow X \leftarrow C \rightarrow Y$ between $Z$ and $Y$. Since we do not know the details about the confounder $C$, we can not use Eq.~\eqref{equ:supp_equ_do} to deconfound $C$. Thus we have to block this path by intervening $X$: 
\begin{equation}
\label{equ:supp_fd3}
    P(Y|do(Z=z))=\sum\nolimits_{x}{P(Y|Z=z,X=x)P(X=x)}.
\end{equation}
At last, by bringing Eq.~\eqref{equ:supp_fd2} and~\eqref{equ:supp_fd3} into Eq.~\eqref{equ:supp_fd1}, we have:
\begin{equation} \label{equ:front_door_supp}
\small
\begin{aligned}
    &P(Y|do(X))\\
    = &\sum\nolimits_{z}P(Z=z|X)\sum\nolimits_{x}P(X=x)[P(Y|Z=z,X=x)],
\end{aligned}
\end{equation}
which is the front-door adjustment given in Eq.~(3) of the submitted manuscript.

\section{Formula Derivations}
\label{sec:supp_nwgm}
Here we show how to use Normalized Weighted Geometric Mean (NWGM) approximation~\cite{xu2015show,srivastava2014dropout} to absorb the sampling into the network for deriving Eq.~(5) in the submitted manuscript. Before introducing NWGM, we first revisit the calculation of a function $y(x)$'s expectation according to the distribution $P(x)$:
\begin{equation}
\label{equ:supp_nwgm1_supp1}
\mathbb{E}_{x}[y(x)]=\sum\nolimits_{x}{y(x)P(x)},
\end{equation}
which is the weighted arithmetic mean of $y(x)$ with $P(x)$ as the weights. 

Correspondingly, the weighted geometric mean (WGM) of $y(x)$ with $P(x)$ as the weights is:
\begin{equation}
\label{equ:supp_nwgm1_supp2}
\text{WGM}(y(x))=\prod\nolimits_{x}y(x)^{P(x)},
\end{equation}
where the weights $P(x)$ are put into the exponential terms. If $y(x)$ is an exponential function that $y(x)=\text{exp}[g(x)]$, we have:
\begin{equation}
\begin{aligned}
\label{equ:supp_nwgm2}
     &\text{WGM}(y(x))=\prod\nolimits_{x}y(x)^{P(x)}\\
     &=\prod\nolimits_{x}\text{exp}[g(x)]^{P(x)} 
    =\prod\nolimits_{x}\text{exp}[g(x)P(x)] \\ &=\text{exp}[\sum\nolimits_{x}g(x)P(x)]=\text{exp}\{\mathbb{E}_{\bm{x}}[g(x)]\},
\end{aligned}
\end{equation}
where the expectation $\mathbb{E}_{x}$ is absorbed into the exponential term. Based on this observation, researchers approximate the expectation of a function as the WGM of this function in the deep network whose last layer is a Softmax layer~\cite{xu2015show,srivastava2014dropout}:
\begin{equation}
\label{equ:supp_nwgm3}
     \mathbb{E}_{x}[y(x)] \approx \text{WGM}(y(x))=\text{exp}\{\mathbb{E}_{x}[g(x)]\},
\end{equation}
where $y(x)=\text{exp}[g(x)]$.

In our case, we treat $P(Y|X,Z)$ (Eq.~(3) of the submitted manuscript) as a predictive function and parameterize it by a network with a Softmax layer as the last layer:
\begin{equation}
\begin{aligned}
\label{equ:supp_nwgm4}
     P(Y|X,Z)=\text{Softmax}[g(X,Z)] \propto \exp[g(X,Z)].
\end{aligned}
\end{equation}

Following Eq.~(3) of the manuscript and Eq.~\eqref{equ:supp_nwgm3}, we have:
\begin{equation}
\begin{aligned}
\label{equ:supp_nwgm5}
& P(Y|do(X))\\
    = &\sum\nolimits_{z}P(Z=z|X)\sum\nolimits_{x}P(X=x)[P(Y|Z=z,X=x)] \\
    = & \mathbb{E}_{[Z|X]}\mathbb{E}_{[X]}[P(Y|Z,X)] 
    \approx  \text{WGM}(P(Y|Z,X)) \\
    \approx & \exp\{[g(\mathbb{E}_{[Z|X]}[Z],\mathbb{E}_{[X]}[X])]\}.
\end{aligned}
\end{equation}
Note that, as in Eq.~\eqref{equ:supp_nwgm4}, $P(Y|Z,X)$ is only proportional to $\exp[g(Z,X)]$ instead of strictly equalling to, we only have $\text{WGM}(P(Y|Z,X)) \approx \exp\{[g(\mathbb{E}_{[Z|X]}[Z],\mathbb{E}_{[X]}[X])]\}$ in Eq.~\eqref{equ:supp_nwgm5} instead of equalling to. Furthermore, to guarantee the sum of $P(Y|do(X))$ to be 1, we use a Softmax layer to normalize these exponential units:
\begin{equation}
\begin{aligned}
\label{equ:supp_nwgm6}
    P(Y|do(X)) \approx \text{Softmax}(g(\mathbb{E}_{[Z|X]}[Z],\mathbb{E}_{[X]}[X])),
\end{aligned}
\end{equation}
where the first part $\mathbb{E}_{[Z|X]}[Z]$ is In-Sample Sampling (IS-Sampling) and the second part $\mathbb{E}_{[X]}[X]$ is CS-Sample Sampling (CS-Sampling). Since the Softmax layer normalizes these exponential terms, this is called the normalized weighted geometric mean (NWGM) approximation. 

In a network, the variables $X$ and $Z$ are represented by the embedding vectors and thus we use $\bm{x}$ and $\bm{z}$ to denote them. Following the convention in attention research where the attended vectors are usually represented in the matrix form, we also pack the estimated IS-Sampling and CS-Sampling vectors to $\hat{\bm{X}}$, $\hat{\bm{Z}}$. In this way, we have:
\begin{equation}
\begin{aligned}
P(Y|do(X)) \approx \text{Softmax}[g(\hat{\bm{Z}},\hat{\bm{X}})], \\
\end{aligned}
\end{equation}
which is given in Eq.~(5) of the submitted manuscript. 

To estimate $\hat{\bm{Z}}$, researchers usually calculate a query set from $X$: $\bm{Q}_I=h(X)$ and use it in the Q-K-V operation. Similarly, to estimate $\hat{\bm{X}}$, we can also calculate a query set as: $\bm{Q}_C=f(X)$ and use it in the Q-K-V operation. In this way, we have Eq.~(5) in the submitted manuscript:
\begin{equation}
\small
\begin{aligned}
P(Y|do(X)) \approx~& \text{Softmax}[g(\hat{\bm{Z}},\hat{\bm{X}})], \\
\textbf{IS-Sampling:} \quad  \hat{\bm{Z}} = & \sum\nolimits_{z} P(Z=z|h(X)) \bm{z}, \\
\textbf{CS-Sampling:} \quad \hat{\bm{X}} = &\sum\nolimits_{x} P(X=x|f(X)) \bm{x}. 
\end{aligned}
\end{equation}
Note that although $P(X)$ in CS-Sampling does not condition on any variable, we still require a query in Q-K-V operation, since without a query, the estimated result will degrade into a fixed single vector for each different input $X$: $\hat{\bm{x}}=\sum\nolimits_{x} P(x) \bm{x}$, where $P(x)$ is the prior probability. We can also treat it as the strategy to increase the representation power of the whole model. 

\section{More Results}
\subsection{Online Captioning Test}
We report the performances of the MS COCO online split in Table~\ref{table:tab_online_supp}. It can be found that our single Transformer+CATT can achieve higher performances than the other state-of-the-art models on this split.
\begin{table}[t]
\begin{center}
\caption{The performances of various single models on the online MS-COCO test server.}
\label{table:tab_online_supp}
\scalebox{0.65}{
\begin{tabular}{l c c c c c c c c c c c}
		\hline
		 Model  & \multicolumn{2}{c}{B@4} &\multicolumn{2}{c}{M} &\multicolumn{2}{c}{R-L} & \multicolumn{2}{c}{C-D}\\ \hline 
		 Metric  &   c5 & c40 & c5 & c40 & c5 & c40 & c5 & c40 \\ \hline
           BUTD~\cite{anderson2018bottom}    & $36.9$ & $68.5$ & $27.6$ & $36.7$ & $57.1$ & $72.4$ & $117.9$& $120.5$  \\ 
           CAVP~\cite{liu2018context}           & $37.9$ & $69.0$ & $28.1$ & $37.0$ & $58.2$ & $73.1$ & $121.6$& $123.8$    \\
           RFNet~\cite{jiang2018recurrent}      & $38.0$ & $69.2$ & $28.2$ & $37.2$ & $58.2$ & $73.1$ & $122.9$& $125.1$    \\
           SGAE~\cite{yang2019auto}    &$37.8$ & $68.7$& $28.1$ & $37.0$ & $58.2$ & $73.1$ & $122.7$ & $125.5$\\ 
           CNM~\cite{yang2019learning}    &$37.9$ & $68.4$& $28.1$ & $36.9$ & $58.3$ & $72.9$ & $123.0$ & $125.3$\\ 
           AoANet${^\dagger}$~\cite{huang2019attention} & $37.3$ & $68.1$ & $28.3$ & $37.2$ & $57.9$ & $72.8$ & $124.0$ & $126.2$ \\ 
           Transformer & $37.9$ & $69.2$ & $28.7$ & $37.7$ & $58.3$ & $73.3$ & $124.1$ & $126.7$ \\ 
           Transformer+CATT & $\bm{38.8}$ & $\bm{70.6}$ & $\bm{28.9}$ & $\bm{38.2}$ & $\bm{58.7}$ & $\bm{73.9}$ & $\bm{126.3}$ & $\bm{128.8}$ \\ 
           
           \hline
\end{tabular}
}
\end{center}
\vspace{-0.3in}
\end{table}

\subsection{More Qualitative Examples}
\begin{figure*}[t]
\centering
\includegraphics[width=0.9\linewidth,trim = 5mm 5mm 5mm 5mm,clip]{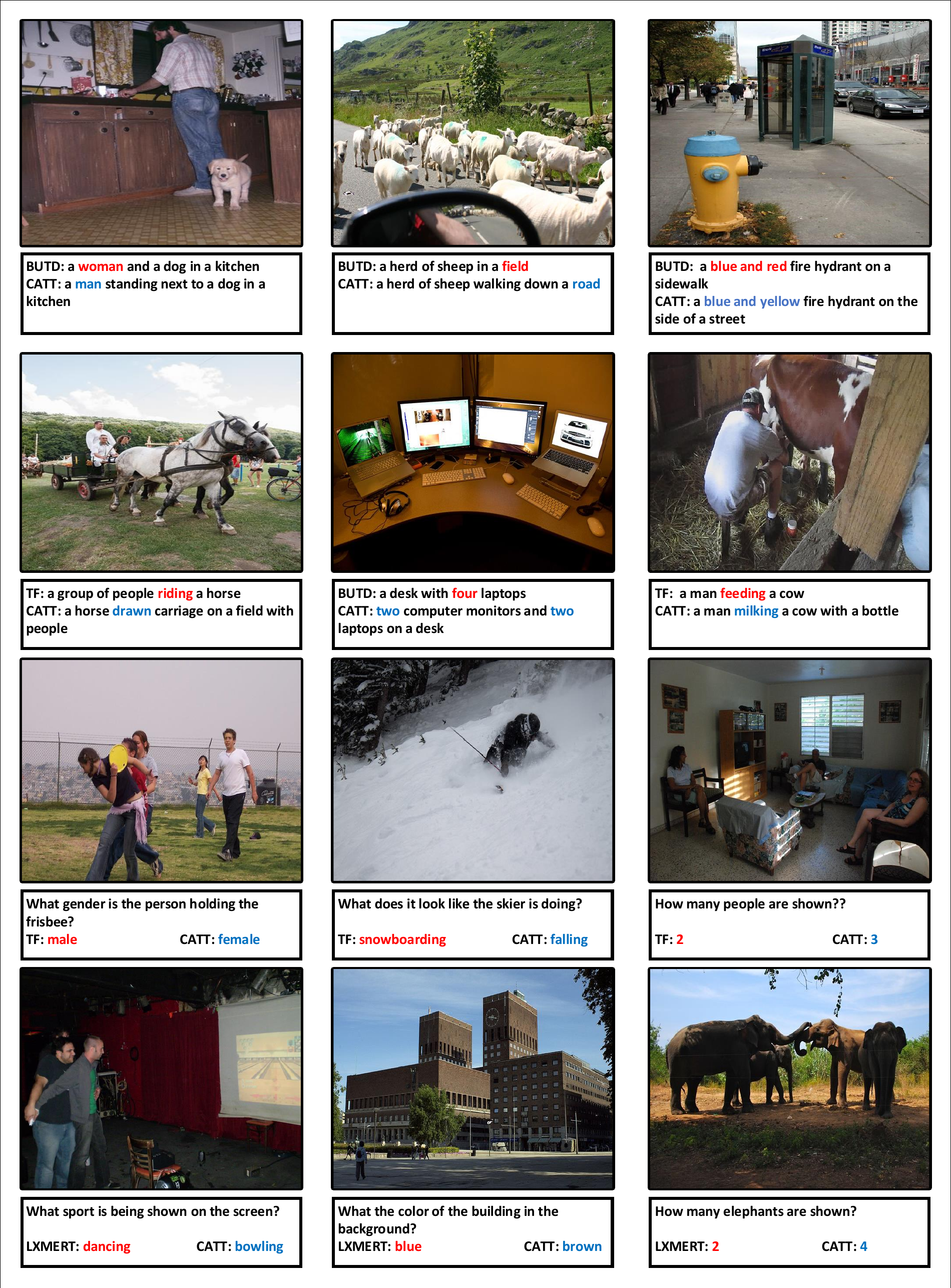}
  \caption{More examples demonstrate that our CATT helps various models confront dataset biases. Red and blue index the incorrect and correct generated captions and answers, respectively.
  }
\label{fig:supp_exp}
\end{figure*}

Figure~\ref{fig:supp_exp} shows more qualitative examples where our CATT helps different models confront the dataset biases. The first two rows show six examples of image captioning and the last two rows show the examples of VQA. For example, in the left example of the first row, after incorporating the CATT module, BUTD~\cite{anderson2018bottom} generates correctly gender of the person without using the spurious correlation between ``woman'' with ``kitchen'' in the dataset.

\section{Implementation Details}
\label{sec:implementation_supp}
\noindent\textbf{BUTD + CATT.}
We deployed this architecture for addressing IC and VQA. In the original BUTD architecture, they only used one attention module and thus we also used one causal attention module as in Figure~\ref{fig:catt}. In this architecture, we set IS-ATT the same as the attention module in BUTD where the probability in Eq.~\eqref{equ:isatt} is calculated as:
\begin{equation} \label{equ:td-att}
\small
\begin{aligned} 
&a_n=\bm{w}^T(\bm{W}_k\bm{k}_n+\bm{W}_q\bm{q}),\\
&\bm{\alpha}=\text{Softmax}(\{a_1,...,a_N\}),
\end{aligned}
\end{equation}
where $\bm{w}$ is a trainable vector and $\bm{W}_k$, $\bm{W}_q$ are two trainable matrices. $\bm{V}_I$, $\bm{K}_I$ were both set to the RoI feature set of the current image and $\bm{q}_I$ was the embedding of the sentence context, \eg, the partially generated caption or the question for IC or VQA, respectively. CS-ATT was set to Eq.~\eqref{equ:csatt}, $\bm{q}_C$ was the same as in IS-ATT and $\bm{V}_C$, $\bm{K}_C$ were both set to the visual global dictionary. This dictionary was initialized by applying K-means over all the RoI features in the training set to get 1000 cluster centres and was updated during the end-to-end training. The RoI object features were extracted by a Faster-RCNN~\cite{ren2015faster} pre-trained on VG as in~\cite{anderson2018bottom}. The hidden size of the LSTM layers was set to 1024.

For the IC model, the cross-entropy loss and the self-critical reward~\cite{rennie2017self} were used to train it 35 and 65 epochs, respectively. We used the Adam optimizer~\cite{kingma2014adam} and initialized the learning rate as $5e^{-4}$ and decayed it by 0.8 every 5 epochs. The batch size was set to 100. For the VQA model, we followed~\cite{teney2018tips,anderson2018bottom} to use the binary cross-entropy loss and applied the AdaDelta optimizer~\cite{zeiler2012adadelta}, which does not require to fix the learning rate, to train it 30 epochs. The batch size was set to 512. 

\noindent\textbf{Transformer + CATT.}
We deployed the architecture in Figure~\ref{fig:transformer} for solving IC and VQA. In this architecture, the Q-K-V operations of all IS-ATT and CS-ATT were implemented by 8-head scaled dot product~\cite{vaswani2017attention}:
\begin{equation} \label{equ:multi-head_supp}
\small
\begin{aligned}
 \textbf{Input:} \quad  &\bm{Q},\bm{K},\bm{V} \\
 \textbf{Prob:} \quad  &\bm{A}_i=\text{Softmax}( \frac{\bm{Q}\bm{W}_i^Q(\bm{K}\bm{W}_i^K)^T}{\sqrt{d}} ) \\
 \textbf{Single-Head}:  \quad  &\bm{H}_i=\bm{A}_i\bm{V}\bm{W}_i^V,\\
 \textbf{Ouput:} \quad & \hat{\bm{V}}= \text{Embed}([\bm{H}_1,...,\bm{H}_8]\bm{W}^H),
\end{aligned}
\end{equation}
where $\bm{W}_i^{*}$ and $\bm{W}^{H}$ are all trainable matrices; $\bm{A}_i$ is the soft attention matrix for the $i$-th head;  $[\cdot]$ denotes the concatenation operation, and Embed$(\cdot)$ means the feed-forward network and the residual operation as in~\cite{vaswani2017attention}. We shared the parameters between IS-ATT and CS-ATT in each CATT to keep the outputs staying in the same feature space. Then compared with the original Transformer, the increments of the trainable parameters only come from the global image and word embedding dictionaries, which were initialized by applying K-means over the RoI and word embeddings of the training set. We set the sizes of both dictionaries to 500 and the hidden size of all the attention modules to 512. The RoI object features were the same as in BUTD+CATT. 

For IC, the training included two processes: we first used the cross-entropy loss and then the self-critical reward to train the captioner 15 and 35 epochs, respectively. The learning rates of two processes were initialized as $5e^{-4}$ and $5e^{-5}$ and both of them decayed by 0.8 every 5 epochs. The Adam optimizer was used and the batch size was set to 10. For VQA, we applied the binary cross-entropy loss and the Adam optimizer to train it 13 epochs. We followed~\cite{yu2019deep} to set the learning rate to $\textbf{min}(2.5te^{-5},1e^{-4})$, where $t$ is the training epoch and after 10 epochs, the learning rate decayed by 0.2 every 2 epochs. The batch size was set to 64.
{\small
\bibliographystyle{ieee_fullname}
\bibliography{main}
}

\end{document}